%% file: root.tex
\documentclass[letterpaper, 10 pt, conference]{ieeeconf}  

\IEEEoverridecommandlockouts                              
\usepackage{amsmath}
\usepackage{amssymb}
\usepackage{textcomp}
\usepackage[utf8]{inputenc}
\usepackage[english]{babel}
\usepackage{graphics}
\usepackage{graphicx}
\usepackage[svgnames]{xcolor}
\usepackage{multirow}
\usepackage{mathtools}
\usepackage{subcaption}
\usepackage{comment}
\usepackage[hidelinks]{hyperref}

\include{defines}

\title{\LARGE \bf
Multi-modal Scene-compliant User Intention Estimation in Navigation
}

\author{Kavindie Katuwandeniya$^{1*}$, Stefan~H.\ Kiss$^{1}$, Lei Shi$^{1}$, and Jaime Valls~Miro$^{1}$%
    \thanks{$^{1}$Robotics Institute, University of Technology Sydney, Australia
        {\tt\scriptsize KavindieHansika.Katuwandeniya@student.uts.edu.au}
    }%
}

\begin{document}

\maketitle
\thispagestyle{empty}
\pagestyle{empty}

\begin{abstract}
A multi-modal framework to generate user intention distributions when operating a mobile vehicle is proposed in this work. The model learns from past observed trajectories and leverages traversability information derived from the visual surroundings to produce a set of future trajectories, suitable to be directly embedded into a perception-action shared control strategy on a mobile agent, or as a safety layer to supervise the prudent operation of the vehicle. We base our solution on a conditional Generative Adversarial Network with Long-Short Term Memory cells to capture trajectory  distributions conditioned on past trajectories, further fused with traversability probabilities derived from visual segmentation with a Convolutional Neural Network. The proposed data-driven framework results in a significant reduction in error of the predicted trajectories (versus the ground truth) from comparable strategies in the literature (e.g. Social-GAN) that fail to account for information other than the agent's past history. Experiments were conducted on a dataset collected with a custom wheelchair model built onto the open-source urban driving simulator CARLA, proving also that the proposed framework can be used with a small, un-annotated dataset. 
\end{abstract}


\section{Motivation and Introduction}
Intention recognition is one of the greatest challenges in intelligent Human Robot Interaction (HRI)~\cite{Burke2004}. As humans, when interacting with one another, an intuitive understanding of the other agents' intentions is learnt allowing better collaborative working conditions. When a user and a machine work in a cooperative fashion, the same principles apply so that tighter shared controls can be exercised and instinctively assist each other in achieving the common goal, bypassing the explicit need for extensive and reliable communication between robot and user~\cite{tahboub2006intelligent}.
Defining user intention is however a challenging proposition. Moreover, it is non-deterministic: given the same situation and user, there could be multiple plausible actions followed through, and they should all be allowed due consideration within a Human-Robot Collaboration (HRC) shared control scheme.

In this work we present a HRC framework for navigational tasks, with the incentive of providing assistance to frail and less able users in driving personal mobility devices (PMDs), such as wheelchairs and scooters. In this context, shared control is perceived differently to other situations where human and robot operate collaboratively on a common workspace, such as assembling pieces together, since actions have a direct impact on the safety and comfort of the rider, and new and unstructured environments are commonplace. 

\begin{figure}[t]
    \centering
    \includegraphics[width=0.6\linewidth]{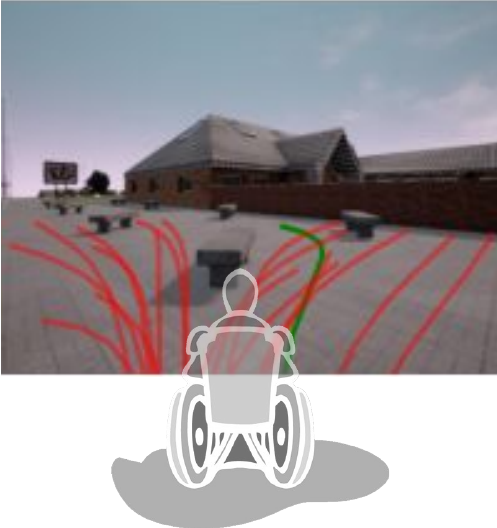}
    \caption{First person view from the on-board camera 
    depicting the \textcolor{Green}{ground truth} trajectory subsequently followed by the user, alongside the equally probable \textcolor{Red}{predicted user intentions} over the same time horizon, projected onto the image.}
    \label{fig:FirstPic}
\end{figure}

We propose a multi-modal data-driven probabilistic approach where the behaviour of a PMD user captured via the kinematics of the vehicle are combined with contextual awareness provided by the scene captured from an on-board monocular camera to estimate the user intention. 
A conditional Generative Adversarial Network (GAN) 
with Long-Short Term Memory (LSTM) encoding and decoding blocks is employed to capture trajectory  distributions conditioned on past trajectories. Rejection sampling is then conducted on future trajectory samples from this distribution to fuse scene traversability information derived from visual segmentation with a Convolutional Neural Network (CNN). The final outcome is a set of equally-probable user intention trajectories,  graphically illustrated in Fig.~\ref{fig:FirstPic}, suitable for integration into a HRC shared control framework for active vehicles, unlike other alternatives where random selection or an oracle are assumed~\cite{Gupta2018, Amirian2019}.

It is rare to find an annotated, considerably large dataset to train generative models of one's domain of work. Our method was developed with the intention of capitalizing on the data available in small, un-annotated datasets.

The paper contributions can be summarised in the development of three components:
\begin{itemize}
    \item a probabilistic approach for multi-modal user intention estimation (Sec.~\ref{sec:Methodology}),
    \item a wheelchair implementation in the simulation environment CARLA~\cite{dosovitskiy2017carla} for safe collection of datasets in realistic urban scenarios (Sec.~\ref{sec:Implementation Dataset}), and
    \item a modular framework able to leverage existing work in the visual segmentation space, suitable to small-scale navigation datasets with no visual annotations (Sec.~\ref{sec:Implementation Dataset} and Sec.~\ref{sec:Implementation_Segmentation}).
\end{itemize}

\begin{figure*}[!tbp] 
    \centering
    \includegraphics[width=\linewidth]{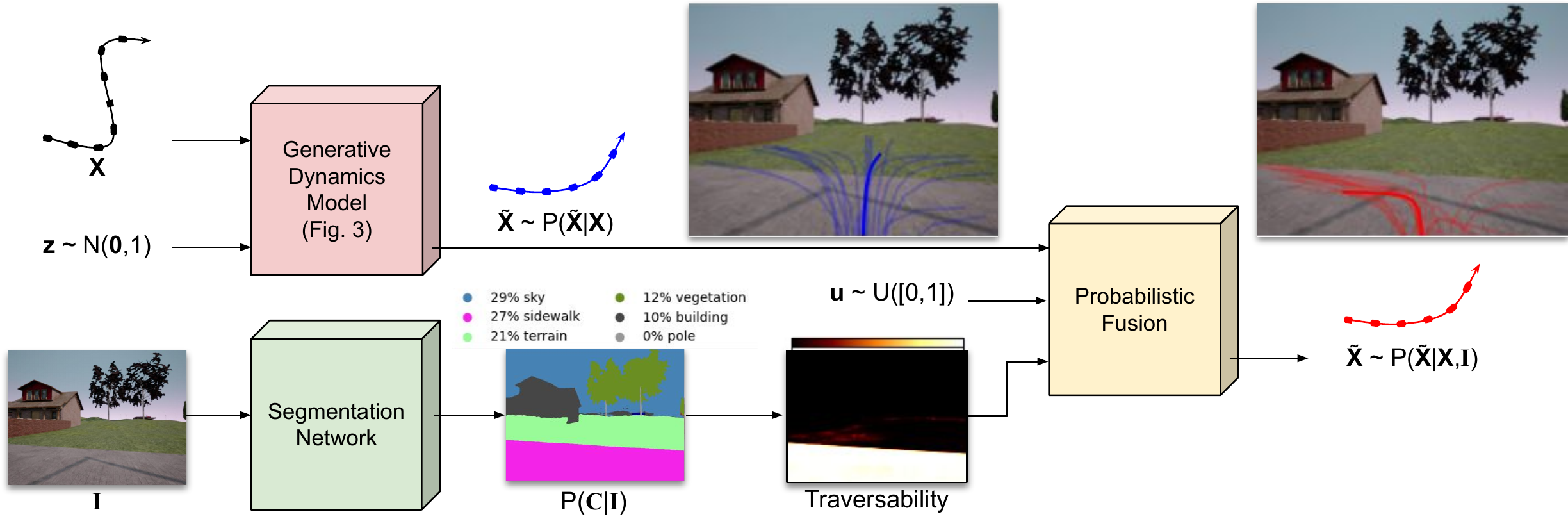}
    \caption{Proposed Architecture with the mean of sampled trajectories given in \textbf{bold}}
    \label{fig:solution}
\end{figure*}
\section{Background}
\subsubsection{User Intention Estimation}
User intention can be interpreted in numerous ways and is not always straightforward to estimate. 
Demeester et al.~\cite{demeester2006bayesian} were the first to break free from the standard definitions of user intention in navigation at that time, which in the context of a robotic wheelchairs were mostly specific assistance algorithms such as ``follow-corridor'', ``avoid-obstacle'', or ``pass through door'', to represent them instead as a set of trajectories each with a goal state. In addition, user intention has been reported in the literature as the immediate controls for the mobility device~\cite{katuwandeniya2020end}, the future trajectory (or control commands of the trajectory) for a predefined time period, or the final goal pose~\cite{narayanan2016semi}. In this work we consider the intended future trajectory to be the user intention. 

\subsubsection{Multi-modal Trajectory Prediction}
Trajectories capture the spatial and temporal relationships of an object. 
Classical methods such as Kalman filters~\cite{kalman1960new} and time-series analysis~\cite{priestley1981spectral} have been used in the past to estimate them from past and current observations. 
However, the complex nature of human behaviours impose the need to disregard hand tailored deterministic models as they fall short in capturing the wide range of plausible future trajectories that represent the user intentions. 
This has given rise to data-driven approaches proven suitable in predicting complex and multi-modal distributions.

Generative models fall under stochastic approaches for data-driven trajectory prediction~\cite{PhanMinh2020CoverNetSets}. Stochastic  approaches  choose  from  multiple  possibilities based on a random sampling. Given  their  flexibility  and proven  ability  to  model complex distributions,  a  generative-model-based  approach  for  trajectory  prediction  was  the  chosen paradigm implemented in our work.

\subsubsection{Generative Models}
Generative models are optimised to capture the underlying data distribution. 
With the advent of deep learning methods, a new set of generative models were introduced: Deep Generative Models (DGMs), including GANs.  

In the context of trajectory prediction, GANs have been extensively utilised in recent literature. \cite{Gupta2018}~was a pioneer in using GANs in the context of identifying and predicting the behaviour of interacting agents with Social-GAN. \cite{Amirian2019}~uses a GAN with an additional information loss to allow for a disentangled representation between the input to the generator and the sampled output. However, both these methods do not include the scene information in their prediction. More recently, \cite{Li2019}~uses a GAN along with scene information to predict the future path of the agents. However, they use a top-down view of the scene which is not possible in our case, or generally for that matter.

Because of instabilities of GANs, predicting trajectories from high dimensional image spaces is difficult. Especially with a small dataset, the chance of over-fitting is high. Thus, it makes sense to use data-driven methods for kinematic-based trajectory prediction and separately perform vision-based scene information integration, where rule-based approaches and pre-trained networks reduce the learning difficulty as they don't need specialisation for the specific hardware or scenario. 

\subsubsection{Segmentation networks}
To develop a solution which does not rely on a large, annotated image dataset, and leverage separate advances in this domain, it is advantageous to make use of pre-trained networks. Thus for the rule-based trajectory selection approach, we chose a pre-trained image segmenting network. For a detailed description of the evolution of deep neural network based image-segmentation methods, the reader is referred to~\cite{sultana2020evolution}.
For the purpose of trajectory selection, it is adequate to understand the different classes in a scene. Thus in this work we focus on semantic segmentation techniques trained on a moving vehicle dataset. 

\section{Methodology}
\label{sec:Methodology}
For this work we define the user intention to be the trajectory the user intends to follow. This definition itself is vague since the trajectory depends on how long it is executed and also the intended destination. Thus it is sensible to define the user intention to be the intended trajectory leading to a point in space at time $t=t+N$ where $N$ is the number of time steps considered into the future. The choice of $N$ could depend on the Field Of View (FOV) of the user or the safe travel distance or time recommended for the mobile platform. 

Mathematically, we denote the position of the wheelchair at discretized time index~$t$ by $\pos_t=(\positionx_t, \positiony_t)$ where $\positionx_t$ and $\positiony_t$ denote the 2-dimensional position at time $t$.
Given the past $M$ positions $\trajpast = \pos_\past$, we wish to predict the next $N$ positions $\trajfuture = \pos_\future$.
We also wish to inform our prediction using scene information: an image~$\image$ captured at time~$t$ from a camera facing forward from the vehicle.
Thus, we are interested in modelling the distribution
\begin{equation}
    \Prob{\trajfuture \given \trajpast, \image}
\\.
\end{equation}

We assume that $\trajpast$ and $\image$ are independent, and also conditionally independent given $\trajfuture$:
\begin{equation}
\begin{aligned}
    \trajpast &\independent \image
\,,\\
    ( \trajpast &\independent \image ) \mid \trajfuture 
\,.
\end{aligned}
\end{equation}
Given an uninformative uniform prior on~$\trajfuture$, these assumptions allow us to calculate the posterior distribution as proportional to the product of the two conditional probabilities:
\begin{equation}
\label{eq:PXXI}
    \Prob{\trajfuture \given \trajpast, \image}
    \propto
        {\Prob{\trajfuture\given\trajpast} \Prob{\trajfuture\given\image}}
\,.
\end{equation}

A conditional GAN is trained to learn the specific user behaviour via the kinematics of the driven vehicle: $\Prob{\trajfuture|\trajpast}$.
A segmentation network is used to incorporate scene information: $\Prob{\trajfuture|\image}$.
These two sources of information are fused probabilistically to produce a generative model. This approach allows the generative model to be scene compliant when trained with only a small dataset.

\subsection{Trajectory Generative Model}
\label{sec:Methodology_Generative Dynamics Model}
\begin{figure}[b]
    \centering
    \includegraphics[width=\linewidth]{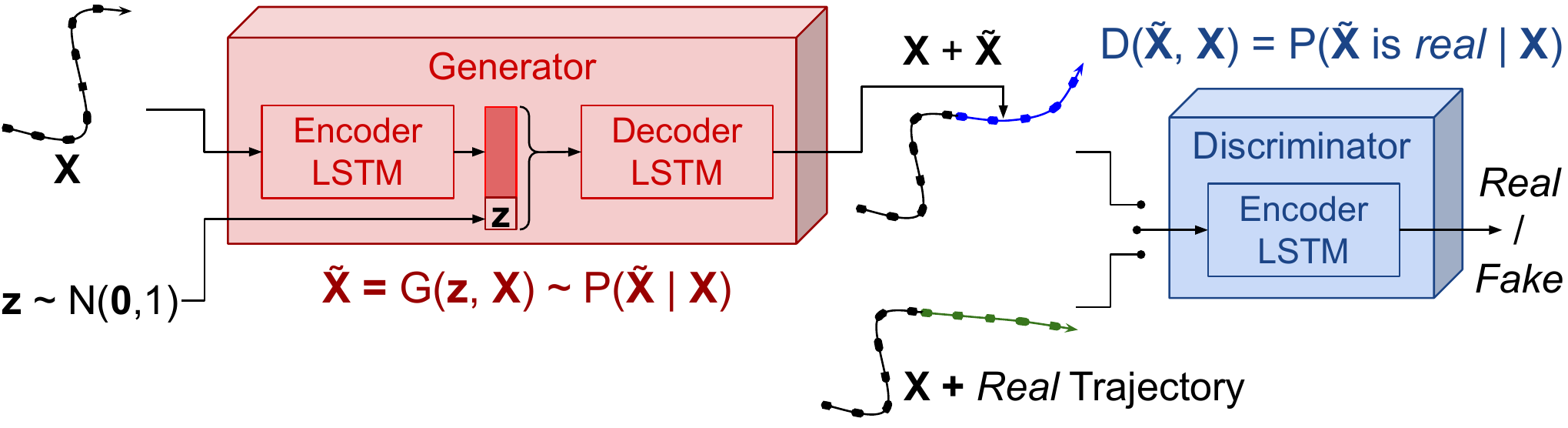}
    \caption{Conditional GAN inspired from Social-GAN~\cite{Gupta2018}}
    \label{fig:socialGAN}
\end{figure}



We  use  a  conditional GAN~\cite{mirza2014conditional} model inspired by~\cite{Gupta2018} to model the user intention from past positions. A GAN framework consists of two adverserial networks: a generator~$G$ and a discriminator~$D$, which are jointly optimized such that the generator learns the underlying data distribution, in our case the user intention as a predicted trajectory distribution. The objective function of a GAN can be written as
\begin{equation}
\begin{split}
    \min_G\max_D V(D,G) =\E_{x\sim \Pr_{data}(x)}[\log D(x)] +\\
    \E_{z \sim \Pr_z(z)}[\log(1 - D(G(z)))]
\end{split}
\end{equation}
where $\Pr_{data}(x)$ is the original data distribution and $\Pr_z(z)$ is a normal distribution from which a noise value is sampled. The discriminator~$D$ is fed with either a real data sample $x$ or a generated sample $G(z)$. 

A conditional GAN is an extended GAN, where the generator and the discriminator are both conditioned on extra information: in our case the observed past trajectory. This framework is shown in Fig.~\ref{fig:socialGAN}.
LSTM blocks are used to capture the temporal relationships of positions. The objective function of the GAN is modified as
\begin{equation}
\begin{split}
    \min_G\max_D V(D,G) =\E_{x\sim \Pr_{data}(x)}[\log D(x|y)] +\\
    \E_{z \sim \Pr_z(z)}[\log(1 - D(G(z|y)))]
\end{split}
\end{equation}
where $y$ conditions both networks $G$ and $D$.

The input for our GAN is the past observed trajectory $\trajpast=\pos_\past$ in the form of consecutive relative positions. In addition to this, a noise vector $\textbf{z}$ is independently sampled from a normal distribution and fed as an input. The output of the generator is a future trajectory in the form of consecutive relative positions, $\trajfuture=\pos_\future$. Different noise inputs result in different predictions, leading to a distribution of possible output trajectories. Resulting trajectories of this network are shown in \textcolor{Blue}{blue} in Fig.~\ref{fig:solution} where $k=20$ samples were generated. 

The data distribution the GAN framework optimizes to learn is $\Prob{\trajfuture\given\trajpast}$ and with different noise values, it outputs different samples from this learnt distribution,
\begin{equation}
    \trajfuture =
    \Gan{\noise\given\trajpast} \sim
    \Prob{\trajfuture\given\trajpast}
\,.
\end{equation}

The main drawback in this framework is that it does not take into account the scene understanding. 
\subsection{Incorporating Scene Information}
\label{sec:Methodology_Incorporating Scene Information}
\begin{figure}[!tbp] 
    \centering
    \includegraphics[width=\linewidth]{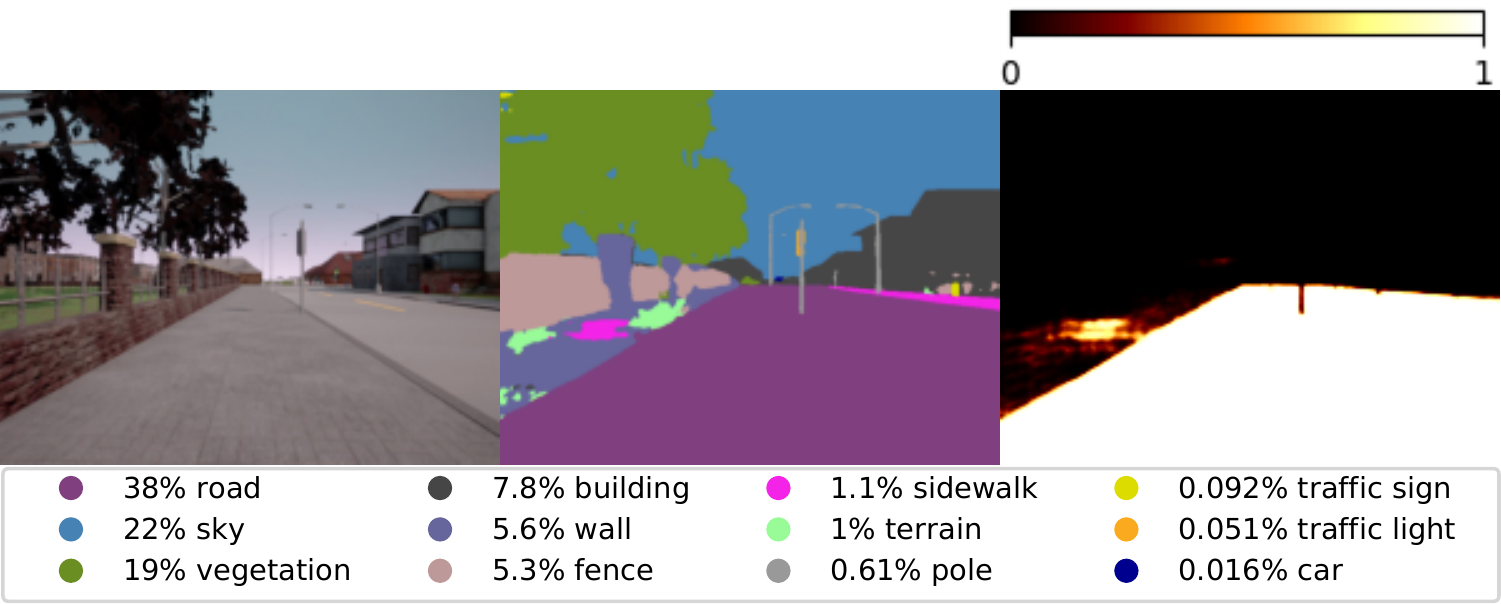}
    \caption{Scene information. \emph{(left)}: original image. \emph{(middle)}: segmented image, coloured by the most likely class. Legend names the 12 most prominent classes. \emph{(right)}: probability of traversability. White indicates the highest probability of 1.}
    \label{fig:segmented-image}
\end{figure}
We integrate the visual information by segmenting the image, generating a probability distribution over pixel-wise classes~$\segmentation$. Gated-SCNN~\cite{takikawa2019gated} was used as the image segmentation network. Further details of the implementation is given under Sec.~\ref{sec:Implementation_Segmentation}. Fig.~\ref{fig:segmented-image} shows the original and segmented images as the \textit{left} and \textit{middle} images respectively.

The segmentation network~$S$ estimates the conditional probability of the scene pixel classes~$\segmentation$ given the image~$I$,
\begin{equation}
    \segmentation \sim 
    \Seg{\image} =
    \Prob{\segmentation\given\image}
\,.
\end{equation}

The $N$ waypoints of a predicted trajectory~$\trajfuture$ are mapped onto the image plane via projective transform~$\projection(\cdot)$.
Vehicle positions are assumed to be 2-dimensional lying on the ground plane ($z=0$) as motion is safely assumed to happen over relatively flat terrain.
The image coordinate~$\mathbf{u}$ of trajectory position~$\pos$ is given by
\begin{equation}
\begin{aligned}
    \begin{bmatrix} \mathbf{u} \\ 1 \end{bmatrix}
    &=
    \mathbf{K}
    \begin{bmatrix} \mathbf{R} & \mathbf{t} \\ \mathbf{0} & 1 \end{bmatrix}^{-1}
    \begin{bmatrix} \pos \\ 0 \\ 1 \end{bmatrix}
\\
    \mathbf{u} &= \projection(\pos)
\,,
\end{aligned}
\end{equation}
where $\mathbf{K}$ is the intrinsic camera matrix and
$\mathbf{R}$ and $\mathbf{t}$ correspond respectively to the 3-dimensional rotation and translation of the camera in world-coordinates.

Out of the 19 class labels that the Gated-SCNN was trained on (more details in Sec.~\ref{sec:Implementation_Segmentation}), ``road'' and ``sidewalk'' were considered traversable and all other class were categorized as non-traversable. Based on this we can assign a probability of traversability to each pixel. An example is shown as the \textit{right} image of Fig.\ref{fig:segmented-image}. By projecting a position~$\pos$ into the image frame, this constraint can be enforced as the conditional probability distribution
\begin{equation}
\begin{aligned}
    \Prob{\pos\given\segmentation}
    &\propto
    \begin{cases}
        1, & \text{if}\ \segmentation(\projection(\pos)) \in \traversable\\
        1, & \text{if}\ \pos\ \text{is at the foot of the robot}\\
        0.5, & \text{if}\ \pos\ \text{is outside the visible area}\\
        0, & \text{otherwise}
    \end{cases}
\end{aligned}
\end{equation}
where $\traversable$ is the set of traversable segmentation classes~$\{\segmentation_\text{road},\segmentation_\text{sidewalk}\}$.
The notation~$\segmentation(\projection(\pos))$ gives the segmentation class for the pixel at position~$\pos$ projected into the image frame. 

Pixels in the segmented image are assumed to be independent, therefore the joint probability of the future trajectory~$\trajfuture$ is computed as the product
\begin{equation}
\begin{aligned}
    \Prob{\trajfuture\given\segmentation}
    &\propto
    \prod_{\pos\in\trajfuture}
    \Prob{\pos\given\segmentation}
\end{aligned}
\end{equation}

Thus, $\Prob{\trajfuture|\image}$ can now be found by,
\begin{equation}
    \Prob{\trajfuture\given\image}
    =
        \sum_\segmentation
        \Prob{\trajfuture\given\segmentation}
        \Prob{\segmentation\given\image}
\,.
\end{equation}


\subsection{Probabilistic Fusion}
We fuse the information about the future trajectory~$\trajfuture$ from the past trajectory~$\trajpast$ and the camera image~$\image$ to refine Eq.~\eqref{eq:PXXI},
\begin{equation}
    \Prob{\trajfuture \given \trajpast, \image}
    \propto
        \underbrace{\Prob{\trajfuture\given\trajpast}}_{\Gan{\noise\given\trajpast}}
        \sum_\segmentation
        \prod_{\pos\in\trajfuture}
        \underbrace{\Prob{\pos\given\segmentation}}_{\substack{\text{traversability}\\\text{constraint}}}
        \underbrace{\Prob{\segmentation\given\image}}_{\Seg{\image}}
\end{equation}

Rejection sampling is used to form a generative model from the posterior distribution~$\Prob{\trajfuture\given\trajpast,\image}$.
Treated as a proposal distribution, the conditional distribution~$\Prob{\trajfuture\given\trajpast}$ is sampled from the GAN conditioned on the past trajectory $\Gan{\noise\given\trajpast}$.
The second conditional probability~$\Prob{\trajfuture\given\image}$ is evaluated using the segmentation of the camera image~$\Seg{\image}$ and considering the traversability constraint~$\Prob{\trajfuture\given\segmentation}$.
The sample is accepted proportional to $\Prob{\trajfuture\given\image}$.
This procedure is repeated until the required number of accepted samples are generated.
As rejection sampling is scale invariant, the proportionality relationships of previous equations are sufficient for evaluation.
The overall framework is illustrated in Fig.~\ref{fig:solution} with the proposed trajectories shown in \textcolor{Red}{red}.

\section{Dataset and Implementation}
\subsection{CARLA Dataset}
\label{sec:Implementation Dataset}
The dataset used for training the GAN framework (Sec.~\ref{sec:Implementation_GAN}) was generated within CARLA~\cite{dosovitskiy2017carla}, an open-source simulation environment for realistic and safe testing of vehicles driving in urban layouts, with humans and vehicles on the road following natural behaviours. A personal mobility device was created and added to the environment for the purpose of this work. A Robotic Operating System (ROS)~\cite{ros} bridge was used to interface CARLA with ROS and an external joystick was used to control the wheelchair. The wheelchair was driven around the environment collecting data for training. 

The dataset consists of six ROS bags each with an average duration of $~10$ minutes. One bag was kept for validating and one for testing, the remaining four were used for training. The training data was organised in $0.5s$ sampling intervals which resulted in $5252$ training instances, $1308$ validating instances and $1261$ testing instances. Even though $5252$ training instance were selected, the environment itself is fairly homogeneous, lacking variation. The wheelchair was driven around city blocks resulting in a relatively small area which produced similar urban imagery in the dataset. 
This lack of diversity in the training instances is believed to be the culprit for the over-fitting we experienced when end-to-end GANs with images as sole input were first tested. Moreover, the data is not annotated and manual labelling would have been a significant time-consuming exercise.

\subsection{GAN Network Implementation}
\label{sec:Implementation_GAN}
The GAN framework was inspired by Social-GAN~\cite{Gupta2018} as in Fig.~\ref{fig:socialGAN}. The integration of LSTM cells allows the network to learn temporal relationships. The generator network consists of an encoder and a decoder network. The encoder encodes the past observed trajectory $\trajpast$ as consecutive relative positions as an encoded vector to which an $8$ dimensional noise vector is added. For this work, $M$ was selected to be $8$ with a $0.5s$ period between time steps. Thus, $\trajpast$ consists of $4s$ of past vehicle trajectory up to the current time step. Each dimension of the noise vector is randomly sampled from a normal distribution independently. 

The hidden layer of the decoder is initialized with the encoded $\trajpast$ and the noise vector. The LSTM cell outputs the consecutive relative future positions as the output from the generator network: $\trajfuture$. As with $\trajpast$, $N$ is selected to be $8$, such that the generator will output a trajectory $4s$ in length starting from the immediate next time step. 

The discriminator takes as input the past trajectory~$\trajpast$ with either the ground truth or the predicted future trajectory~$\trajfuture$, and sends them through LSTM cells to output a scalar value which is then compared against $1$ or $0$: real or fake. As an implementation detail, the labels are smoothed such that $fake \sim \text{Uniform}(0, 0.3)$ and $real \sim \text{Uniform}(0.7, 1)$. $D$ is trained with the binary cross entropy loss of the generated samples and fake labels; and ground truth trajectories and real labels. $G$ is trained using the binary cross entropy loss between the generated samples and real labels. In addition to this variety loss is also considered, where the minimum L2 loss of the generated $k$ samples and the ground truth trajectory is also back-propagated. The Adam optimizer was used to train both networks. The network setup and training procedure followed the Social-GAN framework~\cite{Gupta2018}. The implementation was done in PyTorch~\cite{NEURIPS2019_9015}. 

\subsection{Segmentation Network Implementation}
\label{sec:Implementation_Segmentation}
Gated-SCNN~\cite{takikawa2019gated} was used as the image segmentation network. This model is trainable end-to-end and is trained on the urban Cityscapes dataset~\cite{cordts2016cityscapes} which has 34 semantic labels (of which 19 were used during the training process) from images captured from a road-driving vehicle. Gated-SCNN is a state-of-the-art model, outperforming similar semantic segmentation models considering mean average precision (mAP) as Intersection over Union (IoU) threshold~\cite{sultana2020evolution}. Moreover, the implementation and trained network weights are publicly available and implemented in PyTorch. However, any other semantic segmentation model can be used to replace the Gated-SCNN and the accuracy will change as per the selected model. 

The model consists of 2 streams of networks: a regular stream and a shape stream. The regular stream processes semantic region information using a classical CNN. The shape stream uses low-level feature maps from the regular stream to process the boundary information using Gated Convolution Layers (GCLs). The output of both streams are fed to a fusion module, which outputs semantic regions with clear boundaries. No training of the Gated-SCNN was conducted, the pre-trained model weights were used to semantically segment the image $\image$ at time $t$.

\section{Results and Analysis}


\begin{table}
    \centering
    \begin{tabular}{|r|l|c|c|c|}
        \hline
        &   Selection   &  No scene   & Proposed       & Improvement
        \\
        &               & information &                & 
        \\
        &   & $\Prob{\trajfuture|\trajpast}$ & $\Prob{\trajfuture|\trajpast, \image}$ &
        \\\hline
        \multirow{3}{*}{ADE (m)}
        &   random	&   1.8844	&   1.6094  &   -14.59\%    \\
        &   mean	&   1.6020	&   1.3617  &   -15.00\%    \\
        &   min$_k$ &   0.5406	&   0.5079  &    -6.05\%    \\\hline
        \multirow{3}{*}{FDE (m)}
        &   random	&   4.0259	&   3.3507  &   -16.77\%    \\
        &   mean	&   3.3489	&   2.7988  &   -16.43\%    \\
        &   min$_k$ &   0.7416	&   0.6700  &    -9.65\%    \\\hline
    \end{tabular}
    \caption{Prediction errors with respect to ground truth.}
    \label{tab:result_table}
\end{table}

\begin{figure}[!tbp]
    \includegraphics[width=0.495\columnwidth]{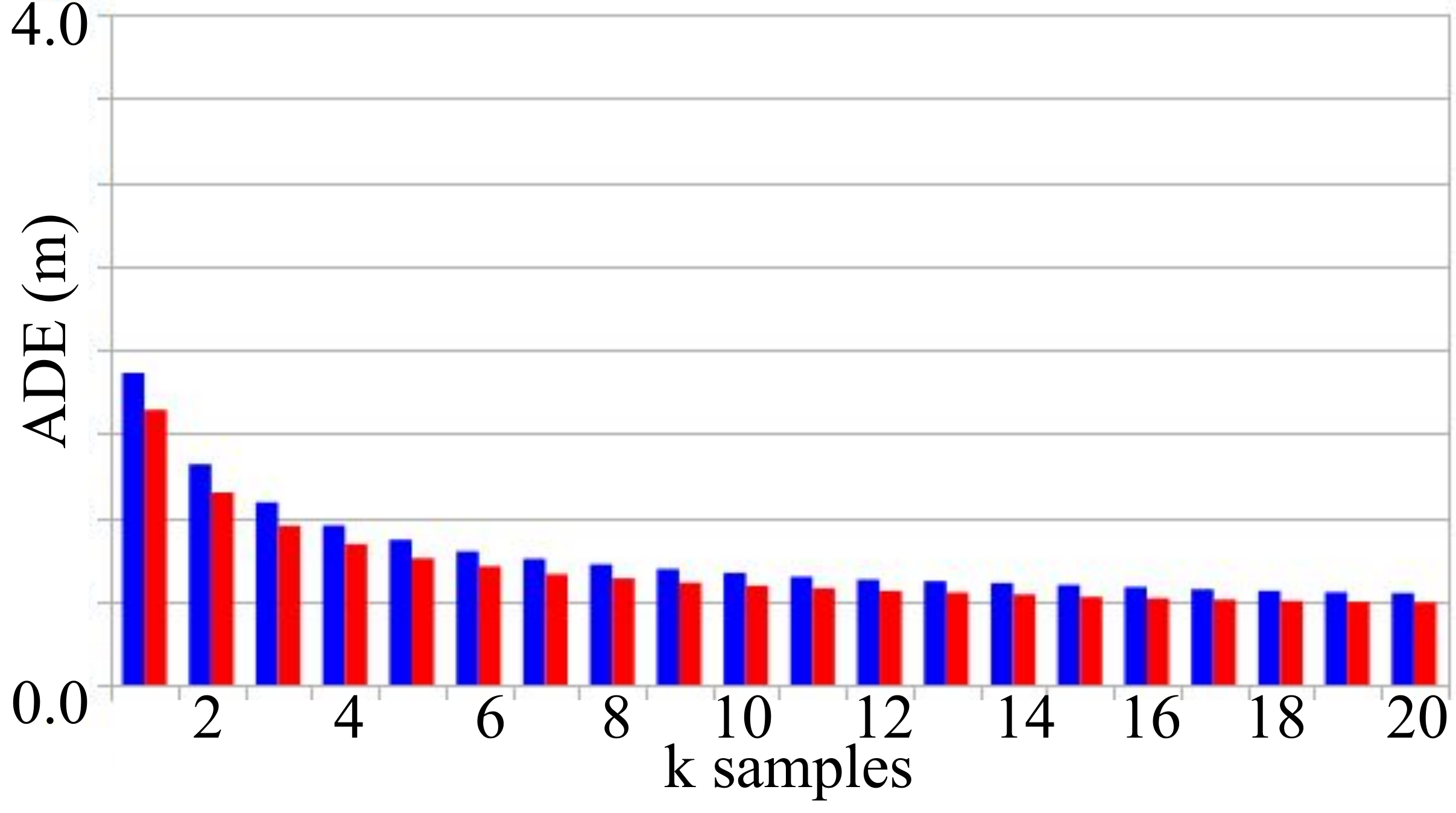}
    \includegraphics[width=0.495\columnwidth]{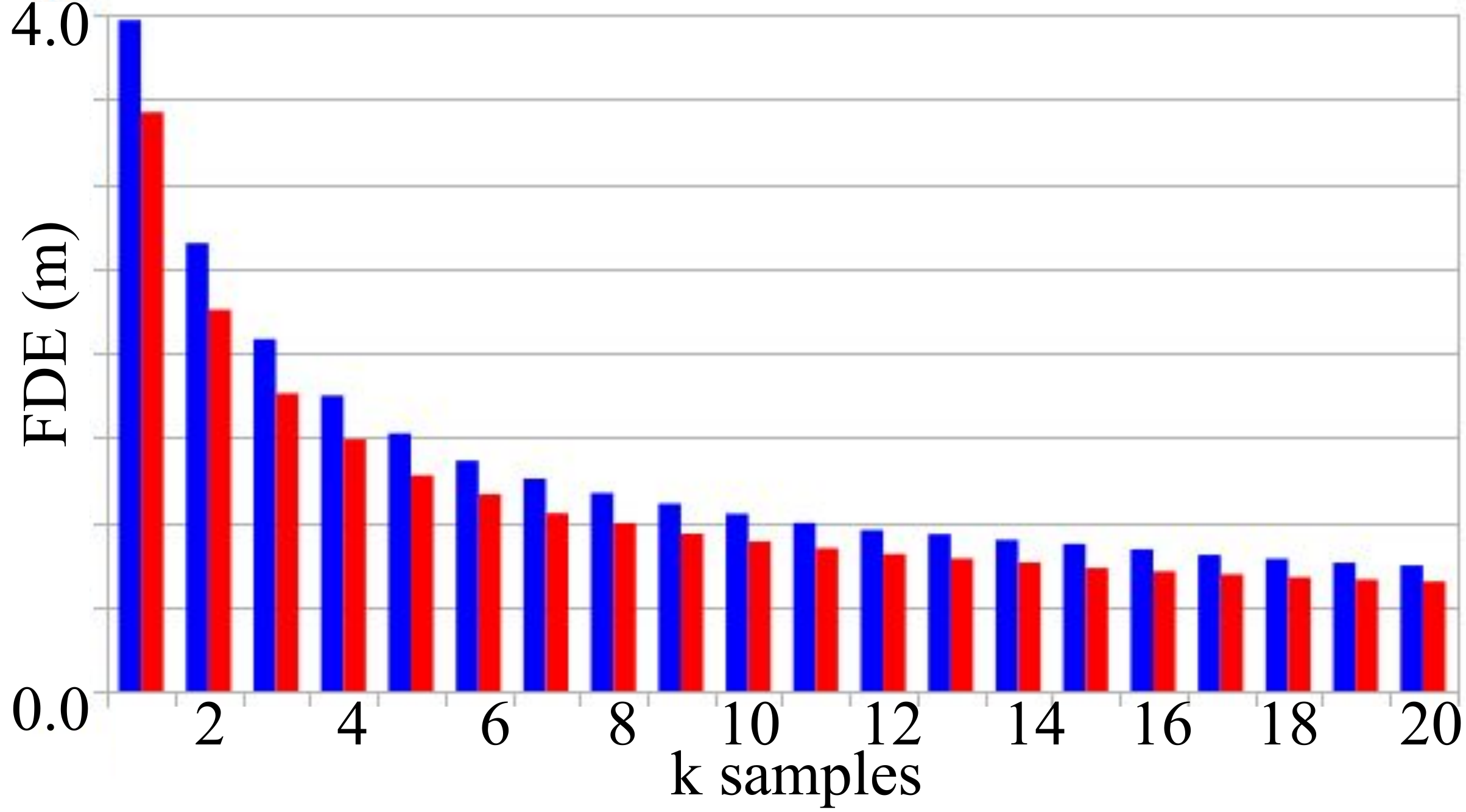}
    \caption{
        Analysis of the min$_k$ metric.
        ADE (left) and FDE (right) of the trajectory closest to the ground truth selected from $k$ samples. \textcolor{blue}{Error with the original set-up ($\Prob{\trajfuture|\trajpast}$)} and \textcolor{red}{error with the proposed framework ($\Prob{\trajfuture|\trajpast, \image})$} shows that our improvement is consistent across the choice of $k$. 
    }
    \label{fig:bestksamples}
    \vspace*{-3mm}
\end{figure}

Average Displacement Error (ADE) and Final Displacement Error (FDE) are common metrics for uni-modal trajectory prediction; ADE gives the mean Euclidean distance along a fixed-duration trajectory while FDE gives the distance error associated to the trajectory endpoints.
According to \cite{Liang2020ThePrediction}, there is no agreed upon metric when it comes to comparing multi-modal prediction frameworks. 
Table~\ref{tab:result_table} shows the improvement of the proposed method with scene understanding to the original (no scene information) Social-GAN framework with different trajectory selection strategies.

As with previous work~\cite{Liang2020ThePrediction,Gupta2018,Amirian2019}, we shall use ADE min$_k$ and FDE min$_k$ as evaluation metrics.
Out of the $k$ samples, the sample which results in the lowest ADE and FDE are used when comparing the results. With the proposed method, there is a $6.05\%$ decrease in the ADE min$_k$ error and $9.65\%$ decrease in the FDE min$_k$ metric as in Table~\ref{tab:result_table}. A comparison figure depicting how the ADE min$_k$ and FDE min$_k$ changes with the number of samples selected is plotted in Fig.~\ref{fig:bestksamples}, reflecting how the decreasing in the error with the proposed method is consistent across any choice of $k$ samples. 

In a real world deployment, if the system is to estimate the user intention and provide the required control assistance based on the estimated intention, one trajectory from the generated $k$ has to be selected. Picking the trajectory closest to the ground truth becomes infeasible. One could either select a random sample (equivalent to min$_1$) or pick the mean of all sampled trajectories.
Our framework shows a consistent decrease in both ADE and FDE under either of these scenarios as per Table~\ref{tab:result_table} under \textit{random} and \textit{mean} selection respectively.

\begin{figure}[t]
    \centering
    \begin{subfigure}{\linewidth}
        \includegraphics[width=\linewidth]{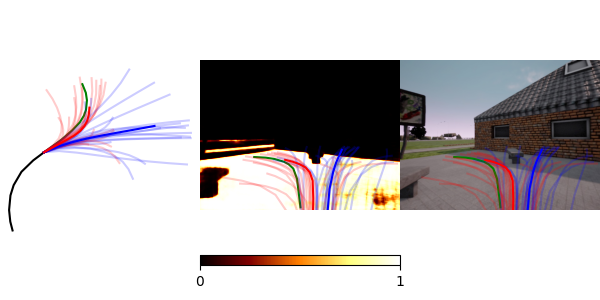}
        \caption{}
        \label{fig:result-example1}
    \end{subfigure} 
    \begin{subfigure}{\linewidth}
        \includegraphics[width=\linewidth]{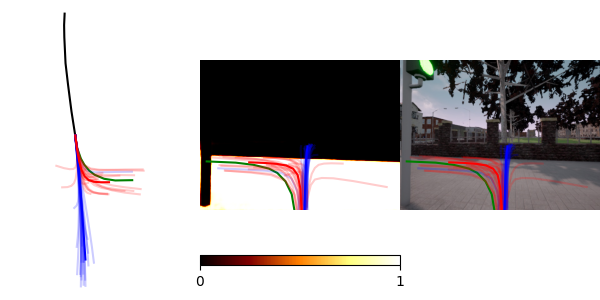}
        \caption{}
        \label{fig:result_example2}
    \end{subfigure}
    \caption{
        Two examples illustrating the proposed methodology.
        \emph{(left)} all the trajectories in the world coordinate frame where 
        \textcolor{Black}{\textbf{observed past trajectory}}, 
        \textcolor{Green}{ground truth trajectory}, 
        \textcolor{Red}{predicted and accepted trajectories (\textbf{mean})},
        \textcolor{Blue}{predicted but rejected trajectories (\textbf{mean})},
        \emph{(middle)} the same trajectories projected onto the segmented image plane.
        \emph{(right)} all the trajectories projected onto the real image plane.
    }
    \label{fig:result-examples}
\end{figure}


Two instances illustrating typical navigation scenarios are shown in Fig.~\ref{fig:result-examples}. If the system was to consider only the kinematics of the assistive device, it would generate the \textcolor{Blue}{blue} trajectories (hereafter the mean of a trajectory distribution will be depicted in \textbf{bold} when projected onto the images). Using the proposed method with consideration for the image information results in (the same number of) equally probable \textcolor{Red}{red} trajectories, compliant with the traversability insights embedded in the image.

Fig.~\ref{fig:result-examples} qualitatively hints at the potential advantage of incorporating visual information into a trajectory generative framework.
Fig.~\ref{fig:result-example1} shows that by rejecting forward- and right-directed trajectories that would have collided with the seat or building, the model predicts more plausible left-directed trajectories, closer to the ground truth.
Fig.~\ref{fig:result_example2} similarly shows a strong increase in left-directed trajectories, but also highlights a weakness of the proposed framework. It can be seen in the projected image that the ground-truth trajectory passes behind a pole, which the segmentation classified as non-traversable. The flawed assumption is that all pixels in the segmented image lie on the ground plane. Various image processing techniques or manipulations to the traversability constraint~$\Prob{\trajfuture\given\segmentation}$ are likely able to correct this shortcoming, and this work is left for future work. 

In this work we have assumed independence for traversability, relying on the traversability generated from the current image to produce $\Prob{\trajfuture\given\image}$. The framework can be improved further by incorporating spatial dependence along the generated trajectory, and utilising more environmental features, which could be learned from historically labelled ground-truth trajectories, e.g. with Conditional Random Fields (CRFs)~\cite{shi2013towards}.


\section{Conclusion}
A framework is developed to estimate user intentions in the space of HRC between a mobility robotic agent and its user. A crucial aspect of this setup is the possibility of having multiple equally-probable estimates for the future trajectory. 
The framework considers the user behaviour through the kinematics of the vehicle and leverages the scene context separately,  probabilistically fusing both information sources to generate the final trajectory distribution. This allows leveraging on separate advances in the image segmentation domain, whilst comparing favourably in its ability to generate trajectories from a small-scale navigation dataset against comparable data-driven schemes in the literature, generating error improvements in the order of $6\%$ or above.   




\section*{Acknowledgement}
The authors would like to thank Janmart Tenedora for his contribution in developing the CARLA simulation environment and the collection of the dataset.


\bibliographystyle{IEEEtran}
\bibliography{references}

\end{document}

%% file: defines.tex
\DeclareMathOperator*\independent{\perp\!\!\!\perp}
\newcommand\E{\mathbb{E}}
\providecommand\given{}
\newcommand\defgiven{\renewcommand\given{\nonscript\:\delimsize|\nonscript\:\mathopen{}}}
\DeclarePairedDelimiterXPP\Prob[1]{\Pr}(){}{\defgiven#1}
\DeclarePairedDelimiterXPP\Expec[1]{\mathbb{E}}[]{}{\defgiven#1}
\DeclarePairedDelimiterXPP\Gan[1]{G}(){}{\renewcommand\given{,}#1}
\DeclarePairedDelimiterXPP\Seg[1]{S}(){}{\renewcommand\given{,}#1}

\newcommand{\pos}{\mathbf{x}}
\newcommand{\positionx}{x}
\newcommand{\positiony}{y}
\newcommand\past{{t+1-M : t}}
\newcommand\future{{t+1 : t+N}}
\newcommand\trajpast{\mathbf{X}}
\newcommand\trajfuture{\widetilde{\mathbf{X}}}
\newcommand\image{\mathbf{I}}
\newcommand\segmentation{\mathbf{C}}
\newcommand\traversable{\segmentation_\text{Traversable}}
\newcommand\projection{H}
\newcommand\noise{\mathbf{z}}

